\pdfoutput=1

\documentclass[11pt]{article}

\usepackage[preprint]{acl}

\usepackage{times}
\usepackage{latexsym}

\usepackage[T1]{fontenc}

\usepackage[utf8]{inputenc}

\usepackage{microtype}

\usepackage{inconsolata}

\usepackage{graphicx}

\usepackage{enumitem}
\usepackage{algorithm}
\usepackage{algpseudocode}
\usepackage{xcolor}

\usepackage{booktabs}
\usepackage{amssymb}
\usepackage{amsmath}
\usepackage{appendix}

\usepackage[switch]{lineno}
\usepackage{tcolorbox}
\usepackage{indentfirst}

\usepackage{multirow}

\usepackage{pifont}

\tcbuselibrary{breakable}
\definecolor{prompt-gray}{HTML}{a7a7a7}
\definecolor{code-function}{HTML}{693da8}
\definecolor{code-syntax}{HTML}{0060b1}
\definecolor{code-constant}{HTML}{d86001}

\title{Unlocking Smarter Device Control: Foresighted Planning with a World Model-Driven Code Execution Approach}

\author{
 \textbf{Xiaoran Yin\textsuperscript{1}\footnotemark[1]},
    \textbf{Xu Luo\textsuperscript{1}\footnotemark[1]},
    \textbf{Hao Wu\textsuperscript{1}},
    \textbf{Lianli Gao\textsuperscript{1}},
    \textbf{Jingkuan Song\textsuperscript{2}\footnotemark[2]}
\\
 \textsuperscript{1}University of Electronic Science and Technology of China,
 \textsuperscript{2}Tongji University
}

\begin{document}
\maketitle
\begin{abstract}
The automatic control of mobile devices is essential for efficiently performing complex tasks that involve multiple sequential steps. However, these tasks pose significant challenges due to the limited environmental information available at each step, primarily through visual observations. As a result, current approaches, which typically rely on reactive policies, focus solely on immediate observations and often lead to suboptimal decision-making. To address this problem, we propose \textbf{Foresighted Planning with World Model-Driven Code Execution (FPWC)}, a framework that prioritizes natural language understanding and structured reasoning to enhance the agent's global understanding of the environment by developing a task-oriented, refinable \emph{world model} at the outset of the task. Foresighted actions are subsequently generated through iterative planning within this world model, executed in the form of executable code. Extensive experiments conducted in simulated environments and on real mobile devices demonstrate that our method outperforms previous approaches, particularly achieving a 44.4\% relative improvement in task success rate compared to the state-of-the-art in the simulated environment.
\end{abstract}
\footnotetext[1]{Equal contributions.}
\footnotetext[2]{Corresponding Author.}
\section{Introduction}
Mobile devices have become central to modern life, facilitating a wide range of activities such as browsing the web, reading news, communicating via email, online chatting, ticket booking, and shopping. As reliance on these devices intensifies, the development of autonomous agents that can seamlessly replicate human interactions to perform such tasks is increasingly crucial. These agents must autonomously interpret textual and visual information from views (i.e., screenshots) and language instructions, navigating the complexity of device interfaces to execute precise actions.

Recent advancements in large language models (LLMs) and vision language models (VLMs)   ~\citep{zhu2023minigpt} provide a promising foundation for developing such agents. When prompted effectively   ~\citep{qin2024mp5}, these models can generate structured, executable actions given task descriptions and observations. Existing works like AppAgent   ~\citep{zhang2023Appagent} and Mobile-Agent   ~\citep{wang2024mobile} have leveraged VLMs to create device-control agents.

However, most existing methods encounter difficulties in making reliable decisions, resulting in a low task success rate. This challenge primarily arises from the complexity of executing tasks on mobile devices, which require a sequence of distinct steps transitioning between views. Each step provides only limited contextual information about the underlying digital environment. Consequently, to accurately predict globally optimal actions, agents must first develop a comprehensive understanding of the environment. Furthermore, they need to implement a forward-looking policy based on this understanding to anticipate the long-term consequences of their actions. In contrast, most prior approaches operate in a ReAct-like action loop   ~\citep{yao2022react}, heavily relying on information from the current observation and a coarse-grained historical context, which leads to myopic decision-making.
\begin{figure*}[t]
  \centering
  \includegraphics[width = 1.0\linewidth]{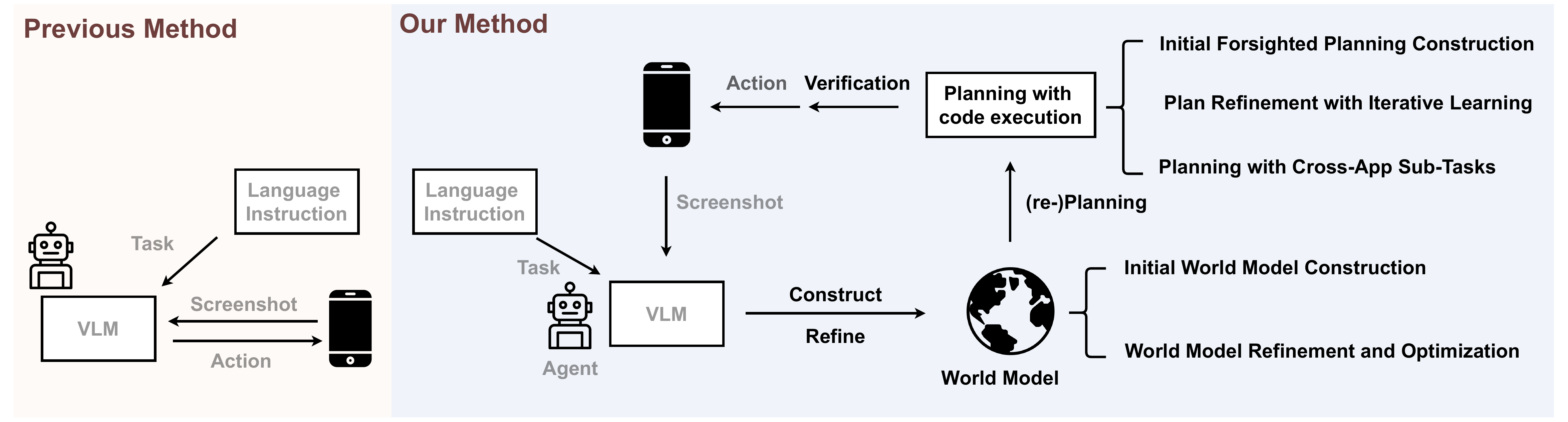}
  \caption{\textbf{Comparison of our proposed framework, Foresighted Planning with World Model-Driven Code Execution (FPWC), with existing approaches.} Unlike prior methods that predominantly rely on reactive actions based solely on immediate observations, our framework leverages a task-specific world model to achieve a more comprehensive understanding of the environment. This world model facilitates iterative planning and the generation of executable code tailored to task completion. Furthermore, both the world model and the planning process are designed to be refined dynamically in an online manner.}
  \label{fig:framework}
\end{figure*}

To overcome these limitations, we propose Foresighted Planning with World Model-Driven Code Execution (FPWC), a novel framework that enables agents to construct a global understanding of the environment they operate in, referred to as the world model   ~\citep{ha2018world} (See Figure \ref{fig:framework}). A world model is essentially a simulation of the environment, such that planning can be done directly within the model, and it enables the agent to predict future states and outcomes without direct interaction with the real environment ~\citep{hafner2019dream}. We represent the world model as a text-based directed graph, where descriptions of views are the nodes and transitions between views are edges. The agent initializes this model given task description, using pre-trained knowledge in VLMs to infer precise, task-relevant contents written as structured texts.  The agent then makes explicit plans by specifying a rule-based policy written as executable code, specifying how an agent should traverse the world model in order to complete the task. As tasks progress, the agent dynamically refines the world model and the planning process in a closed loop, ensuring adaptability and online learning. 

To constrain the complexity of the constructed world model, we strictly limit it to describing a single mobile application that is most pertinent to the task at hand. To adapt our method for tasks necessitating interaction between multiple applications, we incorporate a recursive function within the framework. This function, when called in a plan, generates a sub-agent responsible for executing sub-tasks in other applications that are essential for the current task. The sub-agent mirrors the main agent, possessing its own world models and plans for the sub-task delegated by the main agent, where natural language instructions guide both high-level planning and low-level execution.

In summary, this paper makes the following contributions:

\begin{itemize}[noitemsep,topsep=0pt,partopsep=0pt]
    \item We introduce a novel language-centric task-oriented world model for device-control agents that enhances the agent's global understanding of the environment by representing it as a text-based directed graph, where nodes are linguistic descriptions of views and edges are transitions between them.
    \item We develop a foresighted planning mechanism that allows the agent to generate executable code for task completion by specifying a rule-based policy within the world model, ensuring adaptability and continuous learning through dynamic refinement.
    \item Extensive experiments in both simulated environments and on real mobile devices demonstrate that our approach significantly outperforms existing methods, achieving a 44.4\% relative improvement in task success rate in the simulated environment. These empirical results further yield critical insights into the design and effectiveness of world model-based planning algorithms for mobile device control tasks.
\end{itemize}

\section{Methodology}
To help the device-control agent make informed and foresighted decisions at each step, we endow it with the capability to generate grounded, executable plans based on a self-constructed and dynamically updated world model -- a structured, text-based representation of the app. The following sections provide a detailed overview of our framework FPWC. See Figure \ref{fig:framework} for an overview.
\subsection{Preliminaries}
\textbf{Device-control agents.} We formalize mobile device control as a partially observable Markov decision process (POMDP) with discrete time step $t \in [1, T]$. At each step, given device system state $s_t$, an observation $o_t$ (e.g., screenshots) is generated via an observation function $o_t \sim p(o_t|s_t)$. For a task goal $g$ specified by text, an agent performs an action $a_t$ based on history observations and actions via a \emph{goal-conditioned} policy $a_t \sim \pi(a_t|o_{\leq t}, a_{< t}, g)$, resulting in the next state $s_{t+1}$ via a transition function $s_{t+1}\sim p(s_{t+1}|s_t, a_t)$. A trajectory induced by a policy $\pi$ is defined as $\tau=\{s_t, a_t\}_{t=1}^T\sim p_{\pi}(\tau)$. The goal is to implement a policy $\pi$ that maximizes the expected success rate $\mathrm{E}_{\tau\sim p_{\pi}(\tau)}[\mathrm{Pr}(f(\tau)=1)]$ with the success detector $f(\tau)$.

\noindent\textbf{World models.} Formally, a world model is a simulation of the underlying transition function $p(s_{t+1}|s_t, a_t)$ (and optionally the observation function $p(o_t|s_t)$) of the POMDP ~\citep{ha2018world}. This simulation is achieved through an approximation $p_{\theta}(s_{t+1}|s_t, a_t)$, which is parameterized by $\theta$. Utilizing a learned world model, \emph{planning} can be effectively achieved by optimizing a sequence of actions $\{a_i\}_{i=t}^{T}$ that maximizes the task success rate within the simulated environment constructed by the model: $\max_{\{a_i\}_{i=t}^{T}}\mathrm{E}_{s_{t+1}\sim p_{\theta}(s_{t+1}|s_t, a_t)}[\mathrm{Pr}(f(\tau)=1)]$. Crucially, the planning process is done without interacting with the actual environment.

\subsection{Initial World Model Construction}
A significant challenge in device-control agents is the limited information about the environment that can be obtained from observations. To facilitate a comprehensive understanding of the environment and thereby develop a more anticipatory policy, we construct language-centric world models for device-control agents.

In constructing a world model \( p_{\theta}(s_{t+1}|s_t, a_t) \) as an approximation of the underlying transition function, it is essential to incorporate knowledge of both (1) all linguistically defined states within the environment and (2) all possible transitions between these states. Unlike real-world environments which possess an infinite number of states with continuous changes, the states of mobile devices are finite, and their transitions are discrete. For instance, the settings application on a mobile device typically consists of only a limited number of pages. Therefore, the world model of mobile devices can be effectively represented as a graph $G=(V,K)$, where the set of
 vertices $V$ corresponds to textual descriptions of states (i.e., pages) and the set of edges $K$ represent the possible transitions between these states.
 
In what form should we model the vertices and edges within the graph? Given our objective to furnish the agent with a high-level comprehension of the environment, a detailed graph model is unnecessary. Consequently, we employ natural language descriptions to represent states rather than predicting raw images, and we use high-level action labels, such as ''tap the Wi-Fi button,'' to denote edges. This approach is inspired by the latent states utilized in Dreamer ~\citep{hafner2019dream}. An illustrative example is provided below:

\vspace{0.3em}
\noindent\fbox{\parbox{0.97\linewidth}{\scriptsize{\texttt{{
{\color{prompt-gray}\# Settings app navigation graph\\}
{{\color{code-syntax}Vertices}:}\\
{\hspace*{4mm}{\color{code-function}Name}:} \hspace*{-1mm}``Main page of the Settings app''\\ {\hspace*{4mm}{\color{code-function}Description}:} \hspace*{-1mm}The main page of the Settings App that can be used to navigate to different settings of the phone.\\ \\ {{\color{code-syntax}Edges}:}\\ {\hspace*{4mm}{\color{code-function}Edge}:}\hspace*{-1mm} E({``Main page of the Settings app''}, {``swipe down''}) {{\color{code-syntax}->} ``Main page of the Settings app''}
{\color{prompt-gray}\# Show more settings in the bottom\\}
}}}}}

Each vertex is characterized by a concise name label that encapsulates the state, accompanied by a comprehensive description pertinent to the task. Each edge is defined by a function $\mathrm{E}$, which maps a given vertex and an action to a resulting vertex.  Note that  we do not explicitly model the exact probability distribution of $p_{\theta}(s_{t+1}|s_t, a_t)$. Instead, we enumerate all potential states that may arise from a given state $s_t$ and action $a_t$. The associated uncertainties are addressed through the use of logical statements within code-based planning, as detailed in Section 2.3. This approach is complemented by our functional and structured textual representation of world models, which facilitates seamless integration.

To construct the world model, one potential approach involves manually creating a graph for each specific task. However, this method is labor-intensive and lacks the ability to generalize to novel tasks. Instead, we propose utilizing LLMs to enable the agent to approximate the graph. Specifically, we prompt LLMs to construct the graph given the task instruction before its initial interaction with the environment:
\begin{equation}
    G=\mathrm{LLM}(g),
\end{equation}
where $g$ represents the task goal represented in text form. Given that LLMs have been pretrained on extensive text corpora, we anticipate that they possess the requisite knowledge, although it must be explicitly structured, akin to the chain-of-thought methodology employed in Large Language Models (LLMs) ~\citep{wei2022chain}. Furthermore, we provide an in-context example from a different task to assist the agent in constructing an initial graph that encompasses the essential elements necessary for task completion (see the prompt template in the supplementary material for details). Although this initial model may not be comprehensive, it serves as a foundational framework for subsequent refinement, as elaborated in Section 2.4.

\subsection{Initial Foresighted Planning Construction}

The primary advantage of employing world models lies in their capacity to enable an agent to develop a comprehensive understanding of its environment, thereby facilitating grounded and precise planning. While a straightforward approach to planning involves determining a sequence of actions ~\citep{song2023llm,hafner2019learning}, certain tasks encompass uncertainties or iterative processes that cannot be effectively addressed through a purely sequential method. For instance, in the task of enabling Wi-Fi, the agent must first verify whether Wi-Fi is currently disabled. If it is, the agent should iteratively test various Wi-Fi networks to identify an available connection. To manage such uncertainties, we utilize programming languages, such as Python, which possess the capability to handle logical statements, as the interface for planning. Python's Turing-completeness, structured syntax, and compatibility with vision-language models (VLMs) pretrained on code ~\citep{li2023starcoder,roziere2023code} render it an optimal choice. Given the world model and visual observations, we prompt the VLM to generate a Python function to address the task. An illustrative example of enabling Wi-Fi is provided below:
\vspace{0.3em}
\noindent\fbox{\parbox{0.97\linewidth}{\scriptsize{\texttt{{
{{\color{code-syntax}def} {\color{code-function}plan}():}\\
{\hspace*{4mm}{\color{code-function}E}(}{``Main page of the Settings app''}, {``tap Wi-Fi button''}{)}\\
{\hspace*{4mm}{\color{code-syntax}if not} {\color{code-function}isTRUE}(}{``Wi-Fi button on''}{):}\\ {\hspace*{8mm}{\color{code-function}E}(}{``Wi-Fi (WLAN) settings''}, {``tap the WLAN switch''}{)}\\ {\hspace*{8mm}...}\\ {\hspace*{4mm}{\color{code-syntax}return} {{\color{code-constant}``Task completed''}}
}}}}}}

In our approach, we redefine the edges within the world model, denoted by the callable function $\mathrm{E}$, as abstract actions within the planning process. During the execution of these functions $\mathrm{E}$, the abstract actions (e.g., ''tap Wi-Fi button'') are translated by VLMs into commands executable by mobile devices. Given the limited visual grounding capabilities of VLMs ~\citep{zheng2024gpt}, we incorporate a verification step into the translation process. Specifically, when selecting an element on the views for interaction, we ``zoom in'' by cropping the element and querying the VLMs to confirm whether the element matches their intended target. If the element is not as expected, the VLMs will select an alternative. The construction of the plan is facilitated through meticulously prompted VLMs, with inputs comprising the task instruction, the graph of the world model, and the initial visual observation:
\begin{equation}
    \mathrm{plan}=\mathrm{VLM}(g, G, o_0).
\end{equation}

To address uncertainty, we introduce an additional function, $\mathtt{isTRUE}$, which dynamically verifies specific conditions given current visual observation.
When the graph lacks the necessary information to solve a task, the VLM is capable of synthesizing new vertices and edges, which are marked with an $\mathtt{imagined}$ parameter set to True. Furthermore, an in-context example is included in the prompt to assist the agent in formulating appropriate plans (see the prompt template in the supplementary material for details). To ensure robustness, the agent executes the code within a controlled sandbox environment, where it is dynamically compiled and executed. In the rare event of execution failure, the agent captures the error and recursively refines the plan based on the current context. Our methodology effectively combines the adaptability of world models with the precision of programming, enabling the agent to address both simple and complex tasks effectively.

\begin{algorithm}[t]
\caption{Foresighted Planning with World Model-Driven Code Execution}
\label{alg:main}

\begin{algorithmic}
\Require Language instruction $g$, Initial observation $o_0$, Timestep $T$

\Procedure{Initialization}{}
    \State $G = LLM(g)$\;\Comment{Initialize world model}
    \State $P = VLM(g,G,o_0)$\;\Comment{Initialize plan}
    \State $c = CodeParser.get\textunderscore next\textunderscore line(P)$\;
    \State $t = 0$\; \Comment{Code in plan,refresh step}
    
\EndProcedure

\noindent \textcolor{red}{\textbf{FPWC Loop:}}
    \While{$c$ is not $\varnothing$ and $t<T$}{
        \If{\textcolor{red}{$E(\cdot)$ in $c$}}
            \State $A_{wrong} = \varnothing$ \Comment{Wrong actions set}
            \State $verify = False$ \Comment{Action Verification}
            \While{$verify$ is $False$}
                \State $a_t=VLM(E(\cdot),A_{wrong})$ 
                \State $o_{crop} = Crop(o_t,a_t)$ \Comment{Zoom in}
                \State $verify = VLM(g,o_{crop},E,a_t)$
                \State $A_{wrong} = A_{wrong} \cup \{a_t\}$
            
            \EndWhile
        \State $o_{t+1} = Execute(o_t,a_t)$\\
        \Comment{Refine world model and plan}
        \State $G,P = VLM(g,G,P,o_{t+1},a_t)$ 
    
    \ElsIf{\textcolor{red}{$other\textunderscore app\textunderscore agent(\cdot)$ in $c$}}
    \State $New\textunderscore Agent(subtask,new\textunderscore App\textunderscore name)$
\Else
\State $Execute(c)$  \Comment{Execute normal code}
\EndIf
\State $c = CodeParser.get\_next\_line(P)$\;
\State $t=t+1$\;
\EndWhile
}
\end{algorithmic}
\end{algorithm}

\subsection{World Model and Plan Refinement with Iterative Learning}
An intelligent agent should progressively learn about the environment  ~\citep{zhao2024expel} and revise its knowledge when encountering contradictory observations. In our device-control agent, the initial graph and plan constructed before task execution are only rough estimates and may contain errors. To address this, we enable the agent to refine the world model and plan dynamically based on new observations (see the prompt template in the supplementary material for details):
\begin{equation}
    G_{\mathrm{new}},\mathrm{plan}_{\mathrm{new}} =  \mathrm{VLM}(g,G,\mathrm{plan},o_t,a_{t-1}).
\end{equation}
The action $a_{t-1}$ denotes the most recent action executed by the agent, facilitating the contextualization of the refinement process. During the refinement process of the world model, the agent is permitted to modify the graph structure by adding, removing, or replacing vertices and edges in response to conflicting observations. Regarding the planning process, the agent generates a new, complete plan from the current step onward whenever necessary.

Upon completion of the task, the agent has the option to save the refined graph for future use. This capability allows the agent to leverage prior knowledge, thereby enhancing efficiency over time. The process of self-refinement ensures that the agent continuously adapts, rectifying errors and deepening its understanding with each subsequent task. Further exploration of this mechanism for continual learning on multiple tasks is reserved for future research.

\subsection{Planning with
 Cross-App Tasks}

Mobile devices often require the use of multiple applications to complete certain tasks. For instance, following a user on YouTube may involve first enabling Wi-Fi through the ``Settings'' app if it is turned off, and then returning to the YouTube app to complete the task. Integrating the world models of multiple apps into a single prompt is both challenging and computationally expensive. To address this, our approach focuses on constructing world models for individual apps. To manage cross-app tasks, we introduce a hierarchical agent framework, where a ``parent agent'' can dynamically create a ``child agent'' when a task extends beyond the scope of the current app. This is achieved by incorporating a function, $\mathtt{other\textunderscore App\textunderscore agent}(\mathtt{AppName}, \mathtt{Subtask})$, into the parent agent's plan. When invoked, this function spawns a new computational process to instantiate the child agent, which possesses the same capabilities as the parent agent. The child agent independently constructs its own world model and generates a plan to achieve the specified subtask. During this time, the parent agent's process is temporarily suspended until the child agent completes its task. Once the child agent finishes, control is returned to the parent agent, which resumes its original task. This recursive agent creation mechanism can be invoked multiple times within a single task, as needed. Our full algorithm is summarized in Algorithm \ref{alg:main}.

\section{Experiment}
Through our experiments, we aim to address three pivotal questions that align with the objectives outlined in our study: (1)  How does our approach compare to state-of-the-art device-control methods in terms of metrics such as success rate and efficiency across benchmark tasks?  (2) Can our method effectively tackle tasks that previous approaches fail to solve due to the demand for foresight planning? (3) How well does our agent generalize to tasks on real-world mobile devices, including cross-application scenarios? In order to gain deeper insights, we further evaluate (a) the extent to which individual components contribute to the overall performance of our agent, and (b) the trade-offs between computational efficiency (e.g., latency, resource usage) and task execution performance.

\subsection{Experiment Setup}
To answer the questions above, we conducted experiments primarily utilizing \textbf{MobileAgentBench}, a standardized framework designed for evaluating device-control agents. This benchmark comprises a curated set of 100 tasks spanning 10 open-source applications, including, but not limited to, \textit{Calendar}, \textit{Contacts}, and \textit{Recorder}. By enabling an automated evaluation pipeline, MobileAgentBench ensures reproducibility and mitigates the biases inherent in manual human intervention, thereby producing more reliable and consistent evaluation results.

While MobileAgentBench focuses predominantly on simulation tasks confined to individual applications, we developed an additional \textbf{real-world benchmark} to complement it. This supplementary benchmark encompasses 103 tasks across 9 applications, designed to reflect real-world complexities such as multi-step workflows and physical device interactions (see Appendix table ~\ref{table:real-world bench} for representative examples). By extending the evaluation framework to include physical devices and cross-application tasks, this benchmark provides a broader perspective on the generalizability and adaptability of our proposed approach.

\textbf{Metrics.} 
To comprehensively evaluate the proposed method, a diverse set of evaluation metrics was employed, carefully tailored to the specific characteristics of the \textbf{MobileAgentBench} tasks and the \textbf{real-world} benchmarks. For the \textbf{MobileAgentBench} tasks, we adopted the six metrics originally introduced in the corresponding paper:
\begin{itemize}[noitemsep,topsep=0pt,partopsep=0pt]
    \item \textbf{Success Rate (SR)}: $\mathrm{SR} = \frac{N_{\mathrm{success}}}{M_{\mathrm{tasks}}}$, measuring the proportion of tasks successfully completed.
    \item \textbf{Step-wise Efficiency (SE)}: $\mathrm{SE} = \frac{S_{\mathrm{actual}}}{S_{\mathrm{optimal}}}$, assessing unnecessary/redundant actions (excludes failed tasks).
    \item \textbf{Latency}: Average seconds between consecutive user actions, indicating waiting time.
    \item \textbf{Tokens}: Counts input and output tokens processed by the VLM.
     \item \textbf{False Negative (FN)}: $\mathrm{FN} = \frac{N_{\mathrm{early}}}{N_{\mathrm{failure}}}$, rate of tasks erroneously terminated prematurely.
    \item \textbf{False Positive (FP)}: $\mathrm{FP} = \frac{N_{\mathrm{late}}}{M_{\mathrm{success}}}$,  rate of tasks incorrectly continued after completion.
\end{itemize}

In our real-world evaluations, we concentrate exclusively on the most critical metrics, to assess the adaptability and robustness of agents under varying conditions:
\begin{itemize}[noitemsep,topsep=0pt,partopsep=0pt]
    \item \textbf{Success Rate (SR):} Proportion of tasks successfully completed, following MobileAgentBench's definition. Evaluates task completion under real-world constraints.
    \item \textbf{Completion Rate (CR)}: $\mathrm{CR} = \frac{S_{\mathrm{correct}}}{S_{\mathrm{total}}}$, measuring progress through correctly executed steps versus total required steps. Provides nuanced assessment of partial advancement toward goals.
\end{itemize}

\subsection{Implementation Details}
To empirically evaluate the performance of our device-control agent, we compare with five methods in MobileAgentBench. Additionally, we compared our approach with two representative methods in real-world scenarios. The details of the experiments are provided below (see Appendix for more details).
\begin{itemize}[noitemsep,topsep=0pt,partopsep=0pt]
    \item \textbf{Baseline methods.} The evaluation encompasses the following baselines in MobileAgentBench: AndroidArena   ~\citep{xing2024understanding}, AutoDroid   ~\citep{wen2023empowering}, AppAgent  ~\citep{zhang2023Appagent} (exploration mode disabled), CogAgent   ~\citep{hong2024cogagent} and MobileAgent   ~\citep{wang2024mobile}.
    \item \textbf{Benchmarking Environment.} Both simulated and real experiments are conducted on a Google Pixel 3a emulator running Android 14 operating system, while AutoDroid is tested using Android 9 due to compatibility issues with more recent Android versions. 
    \item \textbf{VLM Standardization.} To ensure fair comparison, all agents utilized GPT-4V as the underlying VLM.
    \item \textbf{Action Space.} Each Agent is allowed to perform five primary actions to simulate device control: $\mathtt{Tap}$, $\mathtt{Long\textunderscore press}$, $\mathtt{Swipe}$, $\mathtt{Text}$, and $\mathtt{Back}$. See  ~\citep{zhang2023Appagent} for details.
\end{itemize}

\subsection{MobileAgentBench Results}
To answer the question of how our approach compares to state-of-the-art (SOTA) device-control methods in terms of success rates and efficiency (Question 1),  we evaluated our agent using the MobileAgentBench benchmark. As shown in Table ~\ref{table:1}, our agent, integrating a world model, planner, and action verifier, achieves the highest  success rate of 39\%, surpassing AutoDroid (27\%), MobileAgent (26\%), and others, thereby demonstrating the efficacy in handling single-app tasks.

In terms of step-wise efficiency, our agent achieves a value of 1.15. Although this is not the lowest among all evaluated methods, it effectively represents the trade-off between efficiency and achieving the highest success rate (SR). This balance underscores the necessity of additional steps to ensure task completion in complex scenarios. Notably, our approach demonstrates a commendable equilibrium between the false negative rate (0.15) and the false positive rate (0.29), indicating a reliable avoidance of premature task termination and a controlled prevention of over-executing actions. In contrast, other methods exhibit either a high false negative rate, such as AutoDroid and CogAgent, or a high false positive rate, as seen in AndroidArena. Our method successfully maintains balanced termination dynamics.

To further explore the trade-offs between computational efficiency and task execution performance (Sub-question (b)), we analyzed the tokens and latency. Our agent processed a higher number of tokens (2120.45) compared to baselines such as CogAgent (579.84) and AutoDroid (963.48). This increased token usage enables dynamic decision-making during task execution, differentiating it from models like AutoDroid, which rely on pre-computed offline strategies and achieve the lowest latency of 4.85 seconds. Despite our method's higher latency (26.13 seconds), it offers greater robustness and adaptability in single-app environments.
\begin{table}
    \centering
    \resizebox{0.5\textwidth}{!}{
    \footnotesize
    \begin{tabular}{l|rrrrrr}
        \toprule
        Agent & SR & SE & Latency (s) & Tokens & FN Rate & FP Rate \\
        \midrule
        AndroidArena* & 0.22 & \textbf{1.13} & 18.61 & 750.47 & \textbf{0.09} & 0.33 \\
        AutoDroid*    & 0.27 & 3.10 & \textbf{4.85} & 963.48 & 0.93 & \textbf{0.01} \\
        AppAgent      & 0.12 & 2.13 & 12.54 & 891.43 & 0.74 & 0.46 \\
        CogAgent      & 0.08 & 2.42 & 6.76 & \textbf{579.84} & 1.00 & 0.04 \\
        MobileAgent   & 0.26 & 1.33 & 15.01 & 1236.88 & 0.19 & 0.31 \\
        \textbf{FPWC} & \textbf{0.39} & 1.15 & 26.13 & 2120.45 & 0.15 & 0.29 \\
        \bottomrule
    \end{tabular}
    }
    \caption{\textbf{Comparison between FPWC and existing methods.} *denotes the agent includes an additional offline exploration phase.}
    \label{table:1}
\end{table}

\subsection{Real-World Results}
e generalization capabilities of our agent in real-world mobile device tasks, including cross-application scenarios (Question 3), we conducted a comparative evaluation against two state-of-the-art baselines: AppAgent~~\citep{zhang2023Appagent} and MobileAgent~~\citep{wang2024mobile}. These baselines were selected due to their strong alignment with real-world usage patterns, particularly their lack of an offline preparation phase, which is critical in mobile device control tasks where immediacy and adaptability are essential.

As shown in Table \ref{table:compare}, our agent achieves a success rate (SR) of 33.0\% and a completion rate (CR) of 62.8\%, compared to AppAgent with an SR of 8.7\% and a CR of 21.7\%, while  MobileAgent achieved an SR of 19.4\%  and a CR of 45.9\%. Furthermore, our agent also demonstrates higher efficiency by completing tasks using the fewest average steps (11.9) compared to AppAgent (14.5) and MobileAgent (13.2).

These results underscore the superior capability of our agent in handling diverse real-world scenarios. By exhibiting higher task success rates, better completion ratios, and optimized step efficiency, our approach sets a new baseline for practical device-control applications. This represents a substantial stride toward more robust and efficient solutions for real-world mobile environments. See Appendix and demo in SM for visualization results for, e.g., cross-app tasks.
\begin{table}[t]
    \centering
    \scalebox{1}{
    \footnotesize
    \begin{tabular}{l|rrr}
        \toprule
        Agent&SR&CR  &Avg. Steps\\
        \midrule
        AppAgent      &8.7\% &21.7\% &14.5 \\
        MobileAgent     &19.4\% & 45.9\%        &13.2\\
        \textbf{FPWC} &\textbf{33.0}\% &\textbf{62.8}\%&\textbf{11.9}  \\ 
        \bottomrule
    \end{tabular}
    }
    \caption{\textbf{Comparison between FPWC and baseline methods on real-world device.} In the absence of standardized evaluation for real-world tasks, the experimental results are assessed by three human experts.}
    \label{table:compare}
\end{table}
\begin{table*}[t]
    \centering
    \footnotesize
    \begin{tabular}{l|ccc|cc}
        \toprule
        Module Combination&Task Comprehension &Long-Horizon Reasoning  &Adaptive Control&SR &Tokens\\
        \midrule
        plain &- &- &- &0.12 &893.58\\
        \ding{172}      &+5 (5) &+7 (7) &+0 (0) &0.24 &1477.01\\
        \ding{172}+\ding{173}     &+1 (6) &+5 (12) &+0 (0) &0.30 &1754.18\\
        \ding{172}+\ding{173}+\ding{174} &+1 (7) &+0 (12) & +4 (4) &0.35 &1979.34\\ 
        \ding{172}+\ding{173}+\ding{174}+\ding{175} &+0 (7)&+0 (12)&+4 (8) &0.39 &2120.45\\
        \bottomrule
    \end{tabular}

    \caption{\textbf{Ablation Study on each component of FPWC.} (\ding{172}:World Model,\ding{173}:Planning,\ding{174}:Self-Refinement,\ding{175}:Self-Verification)}
    \label{table:ablation}
\end{table*}
\subsection{Ablation Study}
To investigate the role of individual components in the overall performance of our agent (Sub-question (a)) and examine its ability to solve tasks requiring foresight planning (Question 2), we performed an ablation study, with results summarized in Table \ref{table:ablation}. Each component is integral to the overall performance of the proposed method. Specifically, the inclusion of the \textbf{world model} significantly enhances decision-making capabilities, demonstrating the pivotal role of environmental context in grounding the agent's actions. Similarly, the \textbf{planning} module contributes to a moderate yet essential improvement, showcasing the utility of anticipatory decision-making for sequential task execution.
The integration of the \textbf{self-refinement} mechanism boosts the success rate from 0.30 to 0.35, underscoring the importance of dynamic plan updates and incremental knowledge adjustment. A marginal improvement is observed with the incorporation of \textbf{self-verification}, which mitigated inaccuracies in visual analysis conducted by GPT-4V. This module ensures task-relevant information is prioritized, thereby delivering more robust and reliable task outcomes.

As expected, resource consumption grows with system complexity due to more frequent model calls and context-rich prompts—consistent with trends observed in LLM-based agents like AutoGPT~\citep{autogpt} and MetaGPT~\citep{hong2023metagpt}, where gains come at computational cost.

To better understand FPWC’s advantage over prior methods, we analyze cases where AppAgent fails but FPWC succeeds. We identify three critical factors: \textbf{Task Comprehension}, \textbf{Long-Horizon Reasoning}, and \textbf{Adaptive Control}.

\textbf{Task Comprehension} refers to the agent’s ability to accurately interpret the semantic intent behind language instructions and map them to the correct UI and functionalities. Traditional reactive agents often depend heavily on immediate visual cues and coarse task goal, which leads to brittle, shallow interpretations of instructions. In contrast, FPWC constructs a task-specific \textbf{world model} (+5) that offers a structured and abstract representation of the environment. As a result, the agent can perform deeper language-to-function grounding—mapping high-level language goals to appropriate UI regions or app modules—even when those are not visible in the current screen. Thus, the world model acts as a semantic scaffold that guides comprehension beyond surface-level patterns.

\textbf{Long-Horizon Reasoning} entails the formulation of coherent multi-step strategies that span multiple views and latent system states. FPWC achieves this by tightly integrating a predictive \textbf{world model} (+7) with a structured, \textbf{code-based planning} (+5) mechanism. The world model encodes transitions and dependencies between UI states as a directed graph, enabling the agent to simulate and reason about future states before acting. Planning is then conducted by generating executable code that operates over this graph, specifying clear, rule-based policies to reach the task goal. Unlike reactive agents—where high-level reasoning (``thought'') is often loosely coupled with concrete actions—FPWC’s use of code ensures that intentions are explicitly and consistently translated into behavior. This eliminates ambiguity, promotes logical consistency, and enforces syntactic structure, resulting in long-range plans that are not only effective but also interpretable and verifiable. 

\textbf{Adaptive Control} refers to the agent’s ability to dynamically adjust its behavior in response to unexpected feedback or execution failures. For example, in the task \textit{``Open About View''}, FPWC initially follows an incorrect path, but upon realizing the mismatch between expected and actual UI, it revises its plan and re-attempts with an alternative strategy. This behavior highlights the role of \textbf{Self-Refinement} (+4), which enables the agent to update its world model and regenerate plans in real time based on new observations. Similarly, in \textit{``Filter media in the gallery and only show images and videos''}, the agent must ensure RAW image formats are excluded. FPWC succeeds by validating UI selections through targeted queries, showcasing the contribution of \textbf{Self-Verification} (+4) in preventing incorrect interactions by confirming element semantics before action execution. Together, these mechanisms promote resilience and precision in complex, state-sensitive tasks.

Our findings highlight that the \textbf{world model} is the key driver of FPWC’s success, enabling accurate instruction grounding and long-horizon reasoning. This underscores that building a structured internal understanding of the environment is essential for robust device-control agent performance.
    


\section{Conclusion}
In this work, we proposed a novel framework for intelligent device control that integrates graph-based world models to enhance task planning and decision-making. By reformulating planning as a code-generation process, our approach enables the iterative refinement of executable Python scripts through a VLM, achieving adaptability and lifelong learning capabilities. To address challenges in spatial localization and fine-grained scene understanding, we introduced a zoom-in self-verification mechanism that improves accuracy in complex environments. Experimental results demonstrated significant performance improvements over existing methods in autonomous device control tasks, validating the effectiveness of our framework.

\section{Limitations}
Despite the promising results, our framework faces several key limitations rooted in the current capabilities of VLMs. First, the precision of interface element recognition remains suboptimal, particularly for small or visually ambiguous components, which can lead to errors in generated code. Second, the VLM's ability to predict long-term outcomes or generate highly specific functions is constrained by its training data and contextual reasoning capacity. These shortcomings may degrade performance in scenarios requiring foresighted planning or precise low-level control. Additionally, the computational cost of token-intensive code generation and verification processes limits real-time applicability. Future work should focus on optimizing model efficiency, enhancing multimodal reasoning, and integrating domain-specific knowledge to overcome these barriers.

\section{Ethical Considerations}

The deployment of autonomous device control systems raises critical ethical concerns that require careful attention. First, the reliance on visual perception and user interface interaction necessitates robust privacy safeguards to prevent unauthorized access to sensitive environments or personal data. Second, potential safety risks arise from misinterpretations of visual inputs or incorrect code execution, which could lead to unintended physical consequences. We emphasize the importance of implementing rigorous validation mechanisms and fail-safe protocols to ensure system reliability. Furthermore, the generalization capability of VLMs introduces biases inherited from their training data, which must be actively mitigated through diverse dataset curation and fairness-aware design. Finally, the accessibility of such frameworks to non-expert users underscores the need for transparent documentation and ethical guidelines to prevent misuse in harmful applications.

\bibliography{custom}

\appendix
\section{Related Work}
\textbf{(Multimodal) Language Agents.} Large language models (LLMs) with instruction following capabilities  ~\citep{longpre2023flan} have enabled the development of language agents  ~\citep{hong2023metagpt} that can reason  ~\citep{yao2024tree}, plan  ~\citep{valmeekam2023planning}, and interact with the real or digital world  ~\citep{liang2023code} using natural language instructions. While text serves as a powerful medium for reasoning and communication, many real-world tasks require multimodal input, particularly vision, which cannot always be effectively translated into text. To address this, vision language models (VLMs)  ~\citep{alayrac2022flamingo} have integrated vision into LLMs, giving language agents ``eyes''. These multimodal agents  ~\citep{xie2024large} can handle tasks requiring fine-grained operations, such as navigation  ~\citep{zhang2024navid}, mobile manipulation  ~\citep{brohan2023rt,stone2023open}, games  ~\citep{wang2024jarvis}, autonomous driving  ~\citep{wang2023drivemlm}, computer operations  ~\citep{hong2024cogagent,koh2024visualwebarena}, and more.

\textbf{Device-control agents.} Research on device-control agents is relatively new. Auto-UI  ~\citep{zhang2023you} uses an VLM trained to directly output actions given views and action histories. AppAgent  ~\citep{zhang2023Appagent} and Mobile-Agent  ~\citep{wang2024mobile} leverage more advanced off-the-shelf VLMs, such as GPT-4v, using a ReAct-style action loop for decision-making. AppAgent combines views and XML files as inputs and includes an optional self-exploration phase to better understand UI elements. Mobile-Agent, on the other hand, relies solely on views but incorporates tools like OCR, Grounding DINO, and CLIP for improved localization. 

While these agents can handle basic tasks, they struggle with issues like shortsightedness, dead loops, and poor performance in cross-App tasks. Our method overcomes these challenges by incorporating a text-based world model that supports action verification, forward-looking planning, and state prediction.

\textbf{World Models.} In the context of reinforcement learning, world models  ~\citep{ha2018recurrent,kaiser2019model}  refer to model-based RL methods  ~\citep{sutton1991dyna} that learn a model to predict the transition dynamics and rewards in the environment, which plays the same role as the environment simulator. As the raw observations are difficult to predict, world models often predict future states in latent representations of an RNN  ~\citep{hafner2019dream,hafner2020mastering}. A learned world model can help for planning  ~\citep{schrittwieser2020mastering} or policy learning by imagination of rollouts  ~\citep{hafner2019dream,janner2019trust}. Some recent works  ~\citep{he2024large,liu2024world} learn world models that directly predict raw image observations by pretraining on a vast number of videos. More recently, there are some works  ~\citep{guan2023leveraging,tang2024worldcoder} utilizing LLMs to construct world models. DECKARD  ~\citep{nottingham2023embodied} builds a text world model of producible items in Minecraft in the form of Python dictionaries, demonstrating how linguistic abstractions improve planning. ~\citep{guan2023leveraging} For example, some studies use LLMs to construct world models in planning domain definition language (PDDL) and Python code, respectively. Compared with these methods, our choice of building world models as a graph using textual interface descriptions and executable code is more suitable for device-control agents.
\section{Experiment Details}
For each views captured from the device, the corresponding XML file is extracted, containing metadata about interactive user interface (UI) elements, such as their types and bounding box coordinates. However, discrepancies may arise where the XML file fails to accurately identify or capture some UI elements. To address this, constraints are introduced: the minimum distance between any two elements must exceed a predefined threshold, and the bounding box of any individual element must not occupy more than a quarter of the screen. Subsequently, the UI elements on the screen are assigned numerical labels in accordance with the following set-of-mark prompting ~\citep{yang2023set}. 

\begin{figure}[t]
  \centering
  \includegraphics[width = 0.5\linewidth]{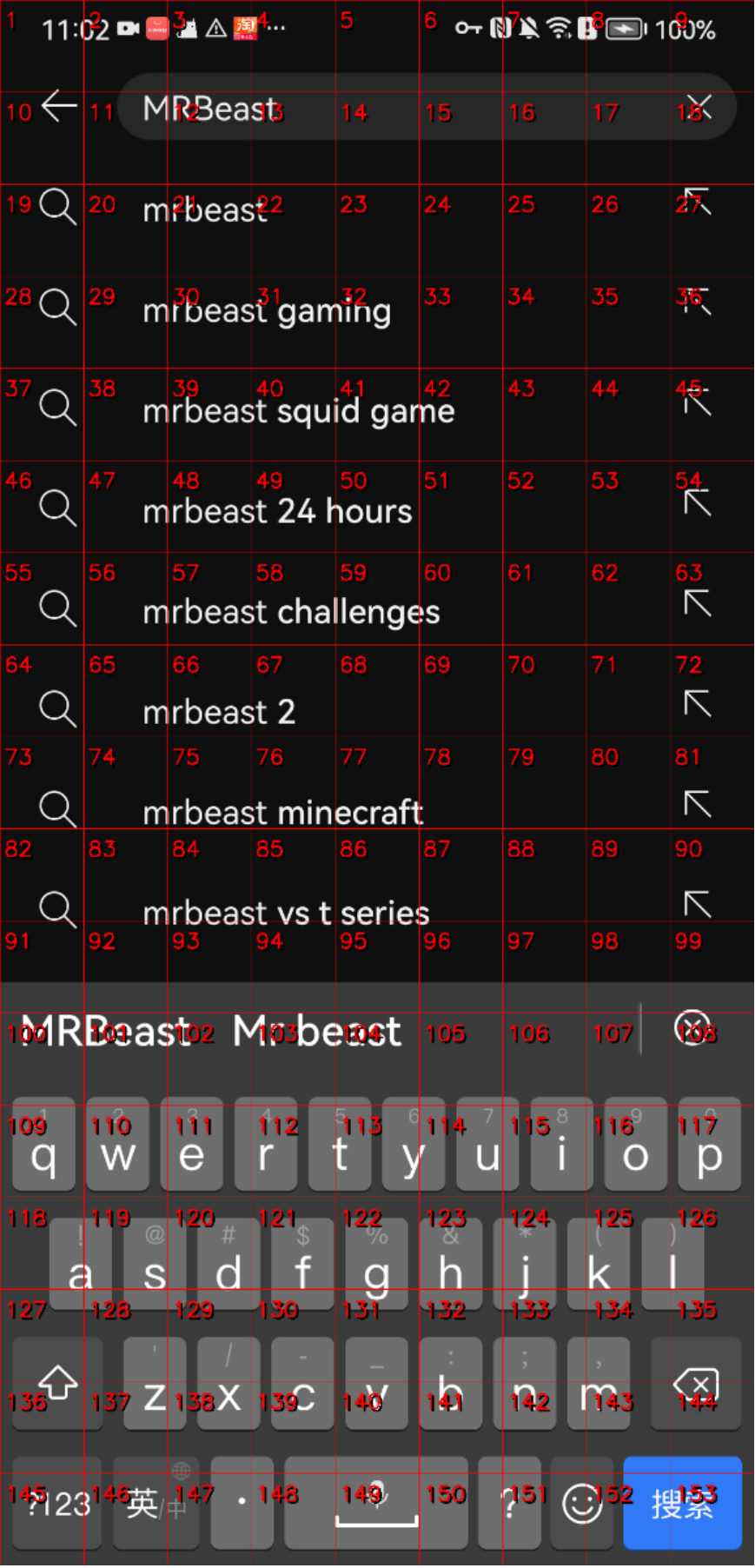}
  \caption{\textbf{Grid Labeling for UI Interaction Verification.} When UI element detection via XML files is unreliable or errors persist, the screen is divided into uniformly sized rectangular grids based on resolution. Each grid cell is assigned a unique identifier, allowing the agent to validate and execute actions accurately by selecting specific regions instead of misidentified elements.}
  \label{fig:grid}
\end{figure}

In case where the XML file omits certain UI elements or the agent repeatedly selects incorrect elements, an alternative approach is employed. The screen is divided into equally sized rectangular grids based on the screen resolution, and each rectangle is assigned a unique number, as Figure \ref{fig:grid} shows.

We use five basic actions:
\begin{itemize}
    \item $\mathtt{Tap}(\mathtt{number}: \mathtt{int})$: this action triggers the tapping operation on the center of the region corresponding to the UI element/grid.
    \item $\mathtt{Long\_press}(\mathtt{number}: \mathtt{int})$: this action triggers the long press operation on the center of the region corresponding to the UI element/grid.
    \item $\mathtt{Swipe}(\mathtt{number}:\mathtt{int}, \mathtt{direction}: \mathtt{str}, \mathtt{dist}: \mathtt{str})$ for XML files and $\mathtt{Swipe}(\mathtt{start\_number}: \mathtt{int}, \mathtt{end\_number}: \mathtt{int})$ for grid: this action triggers the swiping operation. For XML files, the agent should specify 8 directions and three levels of distances. For grids, the agent should specify the start grid and the end grid.
    \item $\mathtt{Text}(\mathtt{text}: \mathtt{str})$: this action directly implements typing of texts.
    \item $\mathtt{BACK}()$: this action is performed on the system level that can make the phone return to the previous level of the current page, usually used to exit an App.
\end{itemize}

The agent is permitted to terminate a task upon successfully executing the entire plan. Additionally, it may conclude the task prematurely, using a designated $\mathtt{FINISH}$ action, if it determines that the objective has already been achieved. For the construction of the world model, we constrain the agent to include only the specified five actions above in the edge, ensuring that the generated plan remains both structured and manageable. 

Once a plan is formulated, it is executed as a Python program. The function `E' is implemented through a sequence of steps, including action selection, validation of the chosen action, and necessary refinement to the graph and plan. If revisions to the plan are required, the current program terminates, and a new execution is initiated. For implementing the $\mathtt{other\_app\_agent}$ function, a new process launches the Python script  with different apps and task configurations, using the $\mathtt{os.system}$ package. The $\mathtt{is\_true}$ function is realized by querying the VLM regarding specific questions related to 
the given image.
\begin{figure}[t]
  \centering
  \includegraphics[width = 1.0\linewidth]{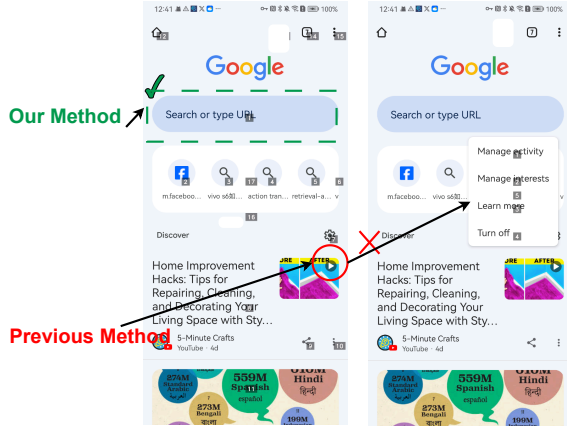}
  \caption{Example highlighting the benefits of employing zoom-in verification prior to executing actions. In this scenario, previous method incorrectly identifies and selects an unintended UI element (Element 7, representing the settings), leading to difficulties in reversing the action. By contrast, our method mitigates such errors by conducting close verification of the elements before proceeding, thereby ensuring accurate action execution.
  }
  \label{fig:verification}
\end{figure}

\section{Graph Complexity Analysis}
To gain clearer insights into the size and complexity of our initial world models, we performed a detailed analysis, the results of which are presented in the Table \ref{complexity}.

\begin{table}[t]
\centering
\caption{Statistics of Initial World Models}
\label{complexity}
\begin{tabular}{lcc}
\toprule
\textbf{Task APP} & \textbf{Vertex Count} & \textbf{Edge Count} \\
\midrule
FileManager   & $6.5 \pm 1.45$  & $16.3 \pm 2.35$ \\
Calculator    & $9.8 \pm 2.12$  & $29.4 \pm 3.02$ \\
Calendar      & $7.5 \pm 1.95$  & $21.0 \pm 2.68$ \\
Contacts      & $6.2 \pm 1.52$  & $15.5 \pm 2.19$ \\
Gallery       & $8.9 \pm 1.79$  & $24.9 \pm 2.86$ \\
Recorder      & $9.5 \pm 2.19$  & $28.5 \pm 2.97$ \\
MusicPlayer   & $7.2 \pm 1.70$  & $20.2 \pm 2.61$ \\
Launcher      & $5.8 \pm 1.22$  & $14.5 \pm 2.02$ \\
Notebook      & $6.1 \pm 1.58$  & $15.3 \pm 2.28$ \\
Messager      & $8.2 \pm 1.87$  & $23.0 \pm 2.76$ \\
YouTube       & $15.5 \pm 2.61$ & $37.8 \pm 3.48$ \\
X             & $14.2 \pm 2.55$ & $34.5 \pm 3.39$ \\
\bottomrule
\end{tabular}
\end{table}

\section{Self-Refinement Analysis}
To more directly evaluate the impact of the self-refinement mechanism, we performed an analysis using our experiment logs.
To clarify how the metrics in our table are calculated, we define the following counts for each app:
\begin{itemize}
    \item $N_{\text{use\_refine}}$: the number of tasks where our agent triggered the refinement mechanism.
    \item $N_{\text{fail\_path}}$: the number of tasks that the baseline agent failed and on which our agent used refinement.
    \item $N_{\text{saved}}$: the number of these ``failing path'' tasks that our agent successfully recovered and completed.
\end{itemize}

\noindent Based on these, the two metrics are calculated as:
\begin{enumerate}
    \item \textbf{Refinement Trigger Rate} = $\frac{N_{\text{use\_refine}}}{10}\times 100\%$
    \item \textbf{Recovery Rate (on Failing Tasks)} = $\frac{N_{\text{saved}}}{N_{\text{fail\_path}}}$
\end{enumerate}
\begin{table*}[t]
\centering
\caption{Refinement Analysis Results}
\label{Refinement}
\begin{tabular}{lcc}
\toprule
\textbf{Task APP} & \textbf{Refinement Trigger Rate} & \textbf{Recovery Rate (on Failing Tasks)} \\
\midrule
FileManager   & 20\% & 0/2 \\
Calculator    & 20\% & 2/2 \\
Calendar      & 20\% & 1/1 \\
Contacts      & 10\% & 0/0 \\
Gallery       & 20\% & 0/1 \\
Recorder      & 20\% & 1/2 \\
MusicPlayer   & 20\% & 0/2 \\
Launcher      & 0\%  & 0/0 \\
Notebook      & 10\% & 0/1 \\
Messager      & 20\% & 1/2 \\
\bottomrule
\end{tabular}
\end{table*}

\noindent This analysis, as summarized in Table ~\ref{Refinement} yields two key insights:
\begin{enumerate}
    \item \textbf{How often?} The refinement mechanism was triggered in approximately 16\% of tasks, primarily in complex scenarios.
    \item \textbf{How well?} When triggered on tasks that the baseline failed, it achieved a recovery rate of approximately 38\% (5 successes out of 13 attempts). This moderate but critical success rate is consistent with our framework's overall performance and accounts for the 5\% absolute gain in success rate.
\end{enumerate}

\section{Per-Step Cost Example}
In this appendix, we provide a comprehensive analysis of latency and token consumption to offer a clearer understanding of the framework's efficiency and scalability.
\begin{itemize}
    \item \textbf{Macro-Level Cost:} At a high level, token consumption serves as a strong proxy for latency. This is because a higher token count for a VLM call directly leads to longer \textbf{inference time}, which is the predominant component of overall latency. Our detailed macro-level token analysis can be found in Table ~\ref{table:ablation}.
    \item \textbf{Micro-Level Example:} For a more intuitive, step-by-step analysis, we annotate Figure~\ref{fig:wifi-example} with the estimated cost for each round. The breakdown for the ``Subscribe to MrBeast'' task is as follows:
\end{itemize}

\begin{table*}[t]
\centering
\caption{Per-Round Token and Latency Breakdown for the ``Subscribe to MrBeast'' Task}
\label{tab:mrbeast_cost}
{ 
\setlength{\tabcolsep}{2pt} 
\begin{tabular}{cccccccccccc}
\toprule
Round 1 & Round 2 & Round 3 & Round 4 & Round 5 & Round 6 & Round 7 & Round 8 & Round 9 & Round 10 & Round 11 & Round 12 \\
\midrule
1387.12 & 1592.45 & 158.67 & 896.33 & 315.78 & 322.10 & 165.23 & 298.99 & 305.21 & 147.01 & 311.05 & 151.89 \\
10.21s  & 12.55s  & 1.82s  & 7.88s  & 3.45s  & 3.51s  & 1.93s  & 3.24s  & 3.37s  & 1.56s  & 3.41s  & 1.62s  \\
\bottomrule
\end{tabular}
} 
\end{table*}
Table~\ref{tab:mrbeast_cost} summarizes this integrated analysis of token and latency estimates, highlighting that complex reasoning steps (planning/refinement) incur the highest costs, whereas verification and execution are relatively economical. This provides a clear view of the framework's cost structure.

\section{Real-World Benchmark}
In Table ~\ref{table:real-world bench}, we show representative examples of the real-world benchmark.
\label{sec:appendix}

\begin{table*}
    \centering
    
    \resizebox{\textwidth}{!}{
    \begin{tabular}{l|l}
        \toprule
        Application&Task description\\
        \midrule
        \multirow{2}{*}{Word}&1. Create a new Word document, write ``Testing123''. Name the document as ``Project1''.\\
& 2. Create a new Word document, write the followings in three lines: (1) Review (2) Edit \\& (3) Submit. Do not include the numbers in the text.\\\midrule
\multirow{2}{*}{Chrome} & 1. Open Chrome, navigate to History and clear all browsing history.\\
& 2. Go to the WIKIPEDIA page for ``Artificial Intelligence'', save it as a PDF, and download it.\\\midrule
\multirow{3}{*}{Gmail} & 1. Send an email to project@example.com with the subject ``AI'' and the body text: ''Hello world!''\\
& 2. Send an email to project@example.com with the subject ''CS'' and the body text: ''Computer Science.'' \\&Attach the first image from the photo album.\\\midrule
\multirow{2}{*}{Maps} & 1. In Maps, search for ``Eiffel Tower'', get driving directions, and start navigation. \\
&2. In Maps ,search for ``Central Park'' and save it to Favorite list.\\\midrule
\multirow{2}{*}{TaoBao} & 1. In TaoBao, add the first item in the cart to the order and proceed to checkout.\\
& 2. Search for ``sunglasses'' in TaoBao and filter results to show items priced between 150 and 200.\\\midrule
\multirow{2}{*}{TikTok} & 1. Change the user's display name to ``John\_Doe'' in TikTok.\\
& 2. In TikTok, follow the creator of the current video being played.\\\midrule
\multirow{2}{*}{WeChat} & 1. Create a new Post in Moments with the caption ``Hello World!'' and attach the first photo from the album.\\
& 2. Send a text message saying ``How are you?'' to File Transfer in WeChat.\\\midrule
\multirow{2}{*}{X} & 1. Follow the user ``@Geoffrey Hinton'' in X.\\
& 2. In X, search for the topic ``Artificial Intelligence'' and like the first tweet in the results.\\\midrule
\multirow{2}{*}{YouTube} & 1. Search for ``AI'' in YouTube and play the first video.\\
& 2. Open the YouTube settings and change the language to ``Spanish.''\\\midrule
\multirow{2}{*}{Multi-App} & 1. Open X, search for the hashtag ``\#MachineLearning'' (with Wi-Fi turned off).\\
& 2. Share a PDF version of the Word document titled ''Project1'' to File Transfer via WeChat.\\
        \bottomrule
    \end{tabular}
    }
    \caption{\textbf{Representative tasks in our real-world benchmark.
    }}
    \label{table:real-world bench}
\end{table*}

\section{Additional Illustrative Examples}
\begin{figure*}[t]
  \centering
  \includegraphics[width = 1.0\linewidth]{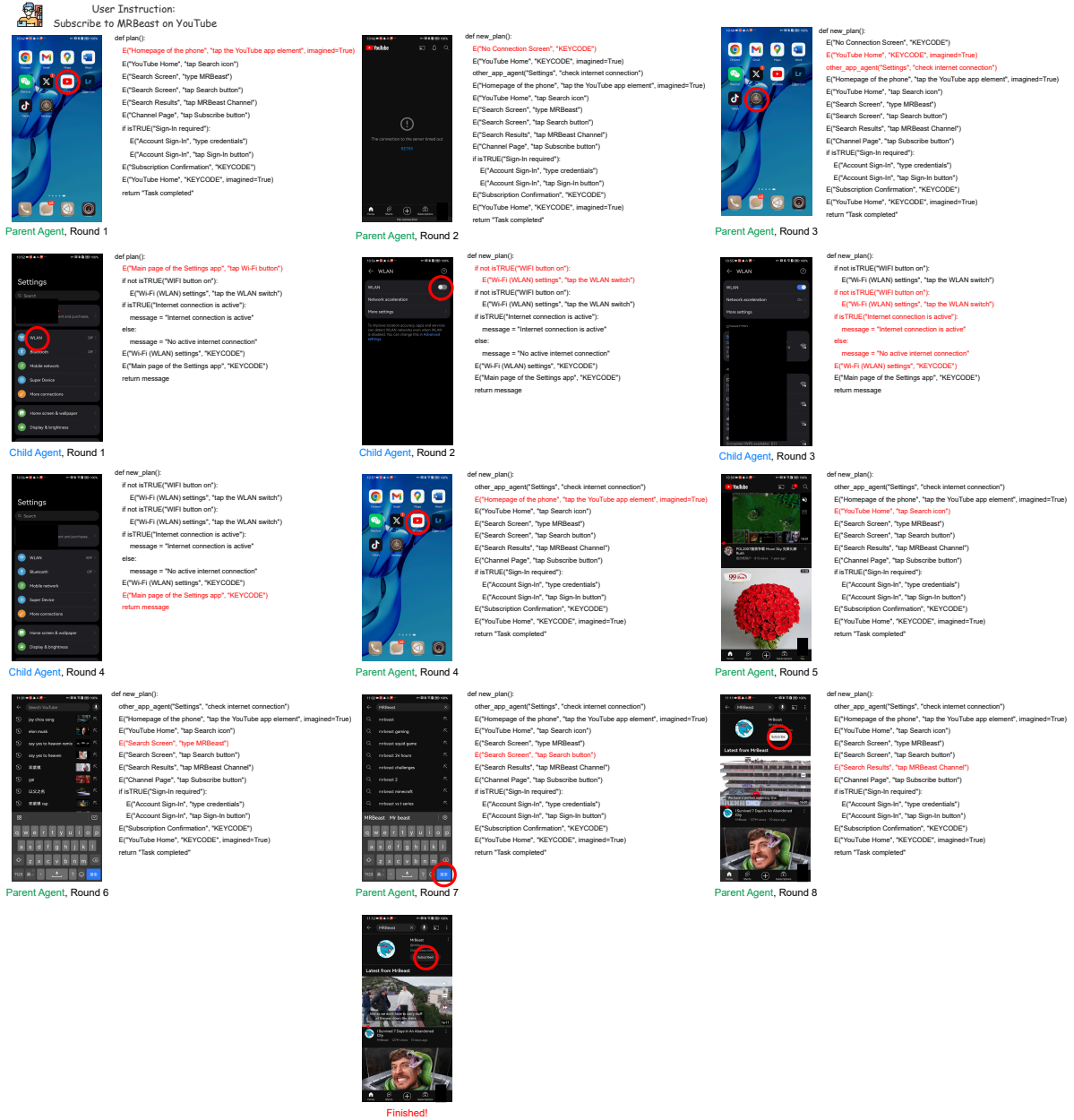}
  \caption{Illustrative example demonstrating the functionality of the proposed method. The task is to subscribe to ``MrBeast''{} on YouTube, with the added challenge that the Wi-Fi is initially disabled—a condition unknown to the agent beforehand. The agent is capable of making long-term decisions while dynamically revising its plan when necessary. Additionally, it can generate a secondary agent to address tasks that require interaction beyond its current application.}
  \label{fig:wifi-example}
\end{figure*}

We provide an illustrative example in Figure \ref{fig:verification} to highlight the benefits of integrating an action verification mechanism prior to the execution of an agent's actions. To further demonstrate the robustness of our approach, Figure \ref{fig:wifi-example} presents a complete example in which the agent successfully performs a complex task: Subscribe to ``MrBeast''{} with Wi-Fi off --- a condition unknown to the agent at the outset. Unlike AppAgent and MobileAgent, both of which fail to accomplish this task, our agent navigates the challenge successfully. A video demonstration of this scenario is included in the supplementary material. Please refer to it if interested.

These examples effectively showcase the feasibility of accomplishing complex tasks under diverse and uncertain conditions. Combined with our results, they highlight the agent's ability to construct a task-oriented world model, anticipate potential challenges, and adapt its actions accordingly. The success in these demonstrative cases underpins the broader applicability of our method and establishes its advantages over previous approaches. Moreover, the detailed documentation and video evidence supplement the findings, offering a transparent basis for validation and replication by the research community. By solving tasks where existing methods falter, our approach emphasizes the practical utility of foresighted policies and supports its contribution to advancing the field of automated device control.

\section{Prompt Template}
This section provides the prompt templates used in our framework.

\onecolumn
\begin{tcolorbox}[
    title={generate\_initial\_world\_model}, 
    breakable,
    fontupper=small
]
You are a graph expert who is trained to generate a graph of a given App on the smartphone.
A task to do is to \textless task\_description\textgreater.
The vertices of the graph is each screen of the App, and the edges are the possible transitions between screens, which are triggered by tapping, swiping or long pressing on UI elements of the screen.
The root of the graph is the initial screen of the App, and the graph should cover as more possible screens and transitions of the App as possible.\\

Output format:

Vertices:\\
Name: \textless Vertex\textgreater{} Description: \textless Description\textgreater{} can-self-act: \textless True or False\textgreater{}
Edges:\\
Edge: E(\textless Vertex\textgreater, action) -\textgreater {} \textless Vertex\textgreater{} \#{} \textless Description\textgreater\\
``can-self-act''{} means that the vertex (screen) can direct to itself, which means there exists actions that can be performed on the screen that does not change the vertex, but change the state of the screen. For example, swipe the main page of the Settings App can show some other choices of settings, but still stay at the main page.
There are only 5 action types: tap, swipe, long\_press, type, BACK (BACK is used to get back to the last level of page, e.g., from the mainpage of an App to the homepage of the phone). Remember that the action type should be put in the first place, for example, ``tap the i-th Wi-Fi network``, ``swipe down/left/right/up``, ``BACK``, ``long\_press the button``, ``type the password``, etc. 
\\
\\
The following shows an example of a partial graph (not complete) of the Settings App:\\

Vertices:\\
Name: ``Main page of the Settings app''{} Description: The main page of the Settings App that can be used to navigate to different settings of the phone. can-self-act: True\\ \\
Name: ``Wi-Fi (WLAN) settings''{} Description: The Wi-Fi settings page that can be used to connect to a Wi-Fi network. can-self-act: True  \\ \\
Name: ``Page connecting to i-th Wi-Fi (WLAN)''{} Description: The page that shows all things needed to connect to i-th Wi-Fi network. can-self-act: True\\ \\
Name: ``Choose Privacy setting of i-th Wi-Fi''{} Description: Use randomized MAC or device MAC. can-self-act: False\\ \\
Name: ``Choose Proxy setting of i-th Wi-Fi''{} Description: Proxy setting of None, Manual or Auto. can-self-act: False\\ \\
Name: ``Choose IP settings of i-th Wi-Fi''{} Description: IP settings of Dynamic or Static. can-self-act: False\\ \\
Name: ``WiFi Connecting''{} Description: The page showing that phone is still trying to connect to the specific WIFI. can-self-act: False\\ \\
Name: ``WiFi password incorrect''{} Description: The page showing that the password of WIFI is incorrect. can-self-act: False\\ \\

Edges:\\
Edge: E(``Main page of the Settings app``, ``BACK``) -\textgreater ``Homepage of the phone''{} \#Back to the phone homepage\\ \\
Edge: E(``Main page of the Settings app``, ``swipe down``) -\textgreater ``Main page of the Settings app''{} \#Show more settings in the bottom\\ \\
Edge: E(``Main page of the Settings app``, ``swipe up``) -\textgreater ``Main page of the Settings app''{} \#Show more settings on the top \\ \\
Edge: E(``Main page of the Settings app``, ``tap Wi-Fi button``) -\textgreater ``Wi-Fi (WLAN) settings''{} \#Open the Wi-Fi setting page\\ \\
Edge: E(``Wi-Fi (WLAN) settings``, ``tap the WLAN button``) -\textgreater ``Wi-Fi (WLAN) settings''{} \#Open Wi-Fi \\ \\
Edge: E(``Wi-Fi (WLAN) settings``, ``tap the WLAN button``) -\textgreater ``Wi-Fi (WLAN) settings''{} \#Close Wi-Fi \\ \\
Edge: E(``Wi-Fi (WLAN) settings``, ``tap the i-th Wi-Fi network``) -\textgreater ``Page connecting to i-th Wi-Fi (WLAN)''{} \#Open the page to connect to i-th Wi-Fi\\ \\
Edge: E(``Page connecting to i-th Wi-Fi (WLAN)``, ``type password``) -\textgreater ``Page connecting to i-th Wi-Fi (WLAN)''{} \#Type the password of the specific Wi-Fi\\ \\
Edge: E(``Page connecting to i-th Wi-Fi (WLAN)``, ``tap the privacy setting``\textgreater -\textgreater ``Choose Privacy setting of i-th Wi-Fi''{} \#Open the page to choose privacy setting of the specific Wi-Fi\\ \\
Edge: E(``Choose Privacy setting of i-th Wi-Fi``, ``tap 'Use randomized MAC' option``) -\textgreater ``Page connecting to i-th Wi-Fi (WLAN)''{} \#Use randomized MAC \\ \\
Edge: E(``Choose Privacy setting of i-th Wi-Fi``, ``tap 'Use device MAC' option``) -\textgreater ``Page connecting to i-th Wi-Fi (WLAN)''{} \#Use device MAC \\ \\
Edge: E(``Choose Privacy setting of i-th Wi-Fi``, ``tap 'CANCEL' button``) -\textgreater ``Page connecting to i-th Wi-Fi (WLAN)''{} \#Cancel the privacy setting \\ \\
Edge: E(``Page connecting to i-th Wi-Fi (WLAN)``, ``tap the proxy setting``\textgreater -\textgreater ``Choose Proxy setting of i-th Wi-Fi''{} \#Open the page to choose proxy setting of the specific Wi-Fi \\ \\
Edge: E(``Choose Proxy setting of i-th Wi-Fi``, ``tap 'None' option``) -\textgreater ``Page connecting to i-th Wi-Fi (WLAN)''{} \#Choose ``None''{} proxy setting \\ \\
Edge: E(``Choose Proxy setting of i-th Wi-Fi``, ``tap 'Manual' option``) -\textgreater ``Page connecting to i-th Wi-Fi (WLAN)''{} \#Choose ``Manual''{} proxy setting\\ \\
Edge: E(``Choose Proxy setting of i-th Wi-Fi``, ``tap 'Auto' option``) -\textgreater ``Page connecting to i-th Wi-Fi (WLAN)''{} \#Choose ``Auto''{} proxy setting\\ \\
Edge: E(``Choose Proxy setting of i-th Wi-Fi``, ``tap 'CANCEL' button``) -\textgreater ``Page connecting to i-th Wi-Fi (WLAN)''{} \#Cancel the proxy setting\\ \\
Edge: E(``Page connecting to i-th Wi-Fi (WLAN)``, ``tap the IP setting``) -\textgreater ``Choose IP settings of i-th Wi-Fi''{} \#Open the page to choose IP setting of the specific Wi-Fi\\ \\
Edge: E(``Choose IP settings of i-th Wi-Fi``, ``tap 'Dynamic' option``) -\textgreater ``Page connecting to i-th Wi-Fi (WLAN)''{} \#Choose ``Dynamic''{} IP setting\\ \\
Edge: E(``Choose IP settings of i-th Wi-Fi``, ``tap 'Static' option``) -\textgreater ``Page connecting to i-th Wi-Fi (WLAN)''{} \#Choose ``Static''{} IP setting\\ \\
Edge: E(``Choose IP settings of i-th Wi-Fi``, ``tap 'CANCEL' button``) -\textgreater ``Page connecting to i-th Wi-Fi (WLAN)''{} \#Cancel the IP setting\\ \\
Edge: E(``Wi-Fi (WLAN) settings``, ``tap 'CANCEL' button``) -\textgreater ``Main page of the Settings app''{} \#Return to the main page \\ \\
Edge: E(``Page connecting to i-th Wi-Fi (WLAN)``, ``tap 'CONNECT' button``) -\textgreater ``WiFi Connecting''{} \#The phone is still trying to connet the specific WIFI\\ \\
Edge: E(``Page connecting to i-th Wi-Fi (WLAN)``, ``tap 'CONNECT' button``) -\textgreater ``WiFi password incorrect''{} \#Incorrect passward for the specific WIFI\\ \\
Edge: E(``Page connecting to i-th Wi-Fi (WLAN)``, ``tap 'CONNECT' button``) -\textgreater ``Wi-Fi (WLAN) settings''{} \#The phone has successfully connected to the specific WIFI\\ \\

Remember only give explanations when the action is self-act or has multiple possible results.\\
The App name is \textless App\_name\textgreater, generate a graph of the App based on your understanding of the APP.
Do not give anything else except the graph in the output.
You should include necessary vertices and edges for completing the task
\end{tcolorbox}

\begin{tcolorbox}[
    title={generate\_initial\_plan}, 
    breakable,
]
You are an intelligent agent trained to complete tasks on a smartphone.
You will generate a Python-like executable plan based on the task description, app graph, and the intial screenshot provided.
Your plan will help navigate through the app screens and complete the task step by step. 
Your output will be used directly as executable code, so follow the required format strictly.\\

To help you reason systematically, follow the steps outlined below:\\
---\\
\#\#\# Context:\\
1. **Input Information**:\\
    - **Task Description**: The task you need to complete is to \textless task\_{description}\textgreater.\\
    - **Graph**: \\
        \hspace{2em}- **Vertices**: \textless Vertices\textgreater 
        \\- **Edges**: \textless Edges\textgreater 
    \\- **Initial Screen**: The smartphone's initial screen is provided as a screenshot Use it as the starting point to create the plan.\\
2. **Graph Details**:
    \\- **Vertices** respresent the app's screens.
    \\- **Edges** define the possible transitions between these screens.Each edge is represented as E(vertex,action),where:
        \\- vertex: The source screen name from which the transition occurs.
        \\- action: The action that leads to the destination screen.
    \\- If you need to imagine new vertices or edges, set the parameter imagined=True in the corresponding function.\\
3. **Available Functions**:
    \\- **E(vertex,action)**: Execute an action to transition to another screen.
    \\- **isTRUE(statement)**: Check if a given statement is true on the current screen.
    \\- **wait()**: Pause execution until the screen changes.
    \\- **other\_app\_function(app\_name,sub\_task)**: Execute high-level tasks in other apps.\\
4. **Action Types**:\\
    There are 5 valid action types:\\
    - 'tap': Taps a specific UI element on the screen.\\
    - 'swipe': Performs a swipe gesture on the screen.\\
    - 'long\_press'': Long presses a specific UI element on the screen.\\
    - 'type': Types a given input into a text field.
    - 'KEYCODE': \\
        - Definition: Simulate hardware key actions like ''Back'' or ''Home''.\\
        - **Use Cases**:\\
            1. Return to the previous screen.\\
            2. Exit an app and return to the phone's main screen.\\
---\\
\#\#\# **Your Task**:\\
1. **Generate a Plan**:\\
    - Your plan must start with the current screen and proceed step by step to complete the task.\\
    - If the current screen is not in the graph, imagine a ''current vertex'' and include 'imagined=True' in any corresponding functions.\\
    - If the task involves actions in another app, you must use the 'other\_app\_function' to complete those actions.Specify the 'app\_name' and 'sub\_task' explicitly.\\
2. **Required Format**:\\
    - **Current Vertex**: Identify the current screen(vertex) based on the provided information. If the screen is imagined, state it explicitly.\\
    - **Plan**: A Python function named 'new\_plan' that implements the steps to complete the task.\\
---\\
\#\#\# **Example Output**:\\
Current vertex:Homepage of the phone\\
Plan:\\
def new\_plan():\\
    \# tap the Settings app element\\
    E(''Homepage of the phone'', ''tap the Settings app element'', imagined = True)\\
    E(''Main page of the Settings app'', ''tap 'Wi-Fi' button'')\\
    \# if the Wi-Fi button is off, turn it on\\
    if not isTRUE(''WIFI button on''):\\
        E(''Wi-Fi (WLAN) settings'', ''tap the WLAN button'')\\

    \# iterate through all Wi-Fi networks on the screen\\
    i = 1\\
    while True:\\
        \# if the i-th Wi-Fi network is out of screen, swipe down to show more Wi-Fi networks\\
        if isTRUE(f''the \{i\}-th Wi-Fi network on the screen is out of screen''):\\
            E(''Wi-Fi (WLAN) settings'', ''swipe down'')\\
            if isTrue(''All Wi-Fi options are the same as before'', compare\_screen = True):\\
                return ''No Wi-Fi with password 57889999 found'' \\
            i = 1\\
        E(''Wi-Fi (WLAN) settings'', f''tap the \{i\}-th Wi-Fi network on the screen'')\\
        E(''Page connecting to i-th Wi-Fi'', ''type password 57889999'')\\
        E(''Page connecting to i-th Wi-Fi'', ''tap 'CONNECT' button'')\\
        if isTRUE(''Wi-Fi connected''):\\
            \# if the Wi-Fi is connected, break the loop\\
            break\\
        elif isTRUE(''Wi-Fi still connecting''):\\
            \# if the Wi-Fi is still connecting, wait for the screen to change\\
            wait()\\
        else:\\
            \# if the password is incorrect, tap the 'CANCEL' button and continue to the next Wi-Fi network\\
            E(''Page connecting to i-th Wi-Fi'', ''tap 'CANCEL' button'')\\
            i += 1\\
    return ''Task completed''\\

\#\#\# **Important Notes**:\\
1. **Strict Output Rules**:\\
    - Your output must strictly follow the required format. Do **NOT** include any description, explanations, or text outside the required format.
    - The plan must be a valid Python-like function with no syntax errors.\\
    - Your output **must start** with 'Current vertex: \textless current\_vertex\textgreater '.\\
    - Always begin the function with 'def new\_plan():'.\\
2. **Error Handling**:\\
    - If the graph is missing necessary information, imagine additional vertices or edges and include 'imagined=True' in the corresponding functions.\\
    - If the task cannot be completed, return an appropriate error message in the plan.\\
3. **Plan Constraints**:\\
    - Always start from the current screen and ensure transitions follow the graph.\\
    - When navigatiing between apps, return to the home screen of the phone using 'KEYCODE' before switching to another app.\\
    - Use 'other\_app\_function' explicitly for actions involving other apps.\\

\end{tcolorbox}

\begin{tcolorbox}[title={refinement}, 
    breakable,]
    You are an intelligent agent trained to complete tasks on a smartphone.
Your role is to analyze the given app graph, the planned code, and two screenshots (before and after an action) to determine whether the plan or the graph needs revision.\\

To help you reason systematically, follow the steps outlined below:\\
---\\
\#\#\# Context:\\
1. **Task**:
    The overall task you need to complete is to \textless task\_description\textgreater .\\
2. **Graph Information**:
    The app you are working on, \textless app\textgreater  , is represented as a directed graph:\\
    - **Vertices**: The screens of the app.\\
        \textless Vertices\textgreater \\
    - **Edges**: The possible transition functions between screens.\\
        \textless Edges\textgreater \\
3. **Plan Code**:\\
    A Python-like plan has been created based on the graph to complete the task. The plan is as follows:\\
    \textless plan\textgreater \\
    The plan may include imagined vertices or edges with the parameter 'imagined=True'.\\
4. **Current Information**:\\
    - The previous screenshot (before the action) corresponds to the vertex: \textless previous\_vertex\textgreater .\\
    - The last action taken corresponds to the function: E(\textless previous\_vertex\textgreater , \textless action\textgreater ).\\
    - The detailed action performed is: \textless detailed\_action\textgreater  (the numeric tag of the UI element).\\
5. **Additional Information**:\\
    - Thoughts: \textless thought\textgreater \\
    - Summary: \textless summary\textgreater \\
    - Action History: \textless act\_history\textgreater \\
    - Repeated Elements: \textless repeated\_doc\textgreater \\
6. **Constraints and Notes**:\\
    - The app graph may need revision based on your observation of the screenshots and the action taken.\\
    - **Action Validity**:\\
        - If the action does not result in a change in the vertex(screen), you must identify it as an ineffective action.\\
        - If the detailed action was wrong (e.g., incorrect numeric tag), you must remind the user to choose a different numeric tag.\\
    - **Plan Revision**:\\
        - If the task is completed but the plan is not, revise the plan to directly return the result.\\
        - If adding **other\_app\_function**, ensure it is included in the **revised plan**.\\
        - Always exit to the home screen of the phone before calling **other\_app\_function**.\\
---\\
\#\#\# **Your Task**:\\
1. Observe the current screenshot (second image).\\
2. Determine the current vertex based on the screenshot.\\
3. If necessary, revise the vertices, edges, and planned code.\\
4. Assess whether the action taken was successful and expected.\\
5. Provide reminders if needed to correct the action or address issues in the plan.\\
---\\
\#\#\# **Output Format**:\\
Your response must strictly follow the format below:\\

Observation of the current screenshot: \textless Describe what you observe in the image\textgreater \\
Thoughts: \textless Explain your reasoning about the graph, plan, and the observed results, and why any revisions are necessary\textgreater \\

Removed vertices:\\
\textless Name: ''\textless vertex\_name\textgreater '' Description: \textless description\textgreater  can-self-act: \textless True/False\textgreater  ... (if no changes, leave blank but keep ''Removed vertices: '' intact)\textgreater 

Added vertices:\\
\textless Name: ''\textless vertex\_name\textgreater '' Description: \textless description\textgreater  can-self-act: \textless True/False\textgreater  ... (if no changes, leave blank but keep ''Added vertices: '' intact)\textgreater 

Removed edges:\\
\textless Edge: E(''\textless start\_vertex\textgreater '', ''\textless action\textgreater '')-\textgreater ''\textless end\_vertex\textgreater '' \#\textless comment\textgreater 
... (if no changes, leave blank but keep ''Removed edges: '' intact)\textgreater 

Added edges:\\
\textless Edge: E(''\textless start
\_vertex\textgreater '', ''\textless action\textgreater '')-\textgreater ''\textless end\_vertex\textgreater '' \#\textless comment\textgreater 
... (if no changes, leave blank but keep ''Added edges: '' intact)\textgreater 
\\
Current vertex: \textless current\_vertex\textgreater 

Added functions for other apps:\\
\textless other\_app\_function(''\textless app\_name\textgreater '', ''\textless sub\_task\textgreater '')
... (if no changes, leave blank but keep ''Added functions for other apps: '' intact)\textgreater 

Successful and expected action: \textless True/False\textgreater 

Ineffective: \textless True/False\textgreater 

Revised plan:
\textless def new\_plan():
    \# Revised code here...
    ...
(if no changes, leave blank but keep ''Revised plan: '' intact)\textgreater 

Remind: \textless Write necessary reminders here; if none, leave blank but keep ''Remind: '' intact\textgreater 

Impact of the action on the element on the task:\\
\textless Describe the effect of the action on the task; this must explicitly connect the transition between the two images and its impact on the task\textgreater \\
---\\

\#\#\# **Example Output**:\\
Removed vertices: \\
Name: ''Main page of the Settings app'' Desciption: Ohhhhhh. can-self-act: True
...
\\
Added vertices: \\
Name: ''Main page of the Settings app'' Desciption: The main page of the Settings App that can be used to navigate to other settings. can-self-act: True
...
\\
The change of edges should be the form like:\\

Removed edges:\\
Edge: E(''Main page of the Settings app'', ''KEYCODE'')-\textgreater  ''Main page of Taobao'' \#Open Taobao
...
\\
Added edges:\\
Edge: E(''Main page of the Settings app'', ''KEYCODE'') -\textgreater  ''Homepage of the phone'' \#Back to the phone homepage
...\\
---\\
\#\#\# **Important Notes**:\\
1. **Removed/Added Vertices and Edges**:\\
    - If any vertices or edges are revised, the removed ones **must exactly match the original graph**, including the comments after ''\#''.\\
    - Added vertices and edges must have clear descriptions and comments.\\
2. **Current Vertex**:\\
    - If the two screenshots differ in content, the vertex should change to reflect the current screenshot.\\
3. **Action Assessment**:\\
    - If the action is ineffective, mark it as True under ''Ineffective''.\\
    - If the action was not successful or expected, mark it as False under ''Successful and expected action''.\\
4. **Revised Plan**:\\
    - If the plan is revised, ensure it begins at the current vertex and includes transitions as required.\\
    - Any add ''other\_app\_function'' **must** be included in the revised plan.\\
5. **Strict Output Format**: All sections must be present, even if left blank.
\end{tcolorbox}

\end{document}